\documentclass{article}
\usepackage{spconf,amsmath,graphicx,hyperref}

\usepackage{amsmath}
\usepackage{multirow}
\usepackage{array}
\usepackage{xcolor}
\graphicspath{{./images/}}
\usepackage{subcaption}
\usepackage{amssymb}
\usepackage{bm}

\usepackage{arydshln} %
\usepackage{graphicx}
\usepackage{booktabs}
\usepackage{tabularx}
\usepackage{enumitem}
\usepackage{seqsplit}

\title{EDITS: Enhancing Dataset Distillation with Implicit Textual Semantics}
\name{Qianxin Xia\textsuperscript{1}, Jiawei Du\textsuperscript{2}, Guoming Lu\textsuperscript{1},  Zhiyong Shu\textsuperscript{1}, Jielei Wang\textsuperscript{1 $\dagger$} \thanks{\textsuperscript{$\dagger$} Corresponding author. This work was supported by the Sichuan Science and Technology Program (Grant No. 2024ZDZX0011, 2026NSFSC1482) and the Jinjiang District 2025 ``Open Bidding for Selecting the Best Candidates" Sci-Tech Project titled ``Edge Computing-Based Lightweight Model for Intelligent Detection of Emergencies in Police Patrol and Its Demonstration Application". This work was also supported by the Shanghai Science and Technology Program (Grant No. 25HB2703100).}}

\address{\textsuperscript{1} University of Electronic Science and Technology of China, Chengdu, China\\
\textsuperscript{2} Centre for Frontier AI Research, Agency for Science, Technology and Research, Singapore
}
\begin{document}
%
\captionsetup[table]{skip=3pt}
\captionsetup[figure]{skip=4pt}
\maketitle

\begin{abstract}
Dataset distillation aims to synthesize a compact dataset from the original large-scale one, enabling highly efficient learning while preserving competitive model performance. However, traditional techniques primarily capture low-level visual features, neglecting the high-level semantic and structural information inherent in images. In this paper, we propose \textbf{EDITS}, a novel framework that exploits the implicit textual semantics within the image data to achieve enhanced distillation. First, external texts generated by a Vision Language Model (VLM) are fused with image features through a Global Semantic Query module, forming the prior clustered buffer. Local Semantic Awareness then selects representative samples from the buffer to construct image and text prototypes, with the latter produced by guiding a Large Language Model (LLM) with meticulously crafted prompt. Ultimately, Dual Prototype Guidance strategy generates the final synthetic dataset through a diffusion model. Extensive experiments confirm the effectiveness of our method. Source code is available in: \url{https://github.com/einsteinxia/EDITS}. 

\end{abstract}

\begin{figure}[t]
  \centering
  \centerline{\includegraphics[width=8.35cm]{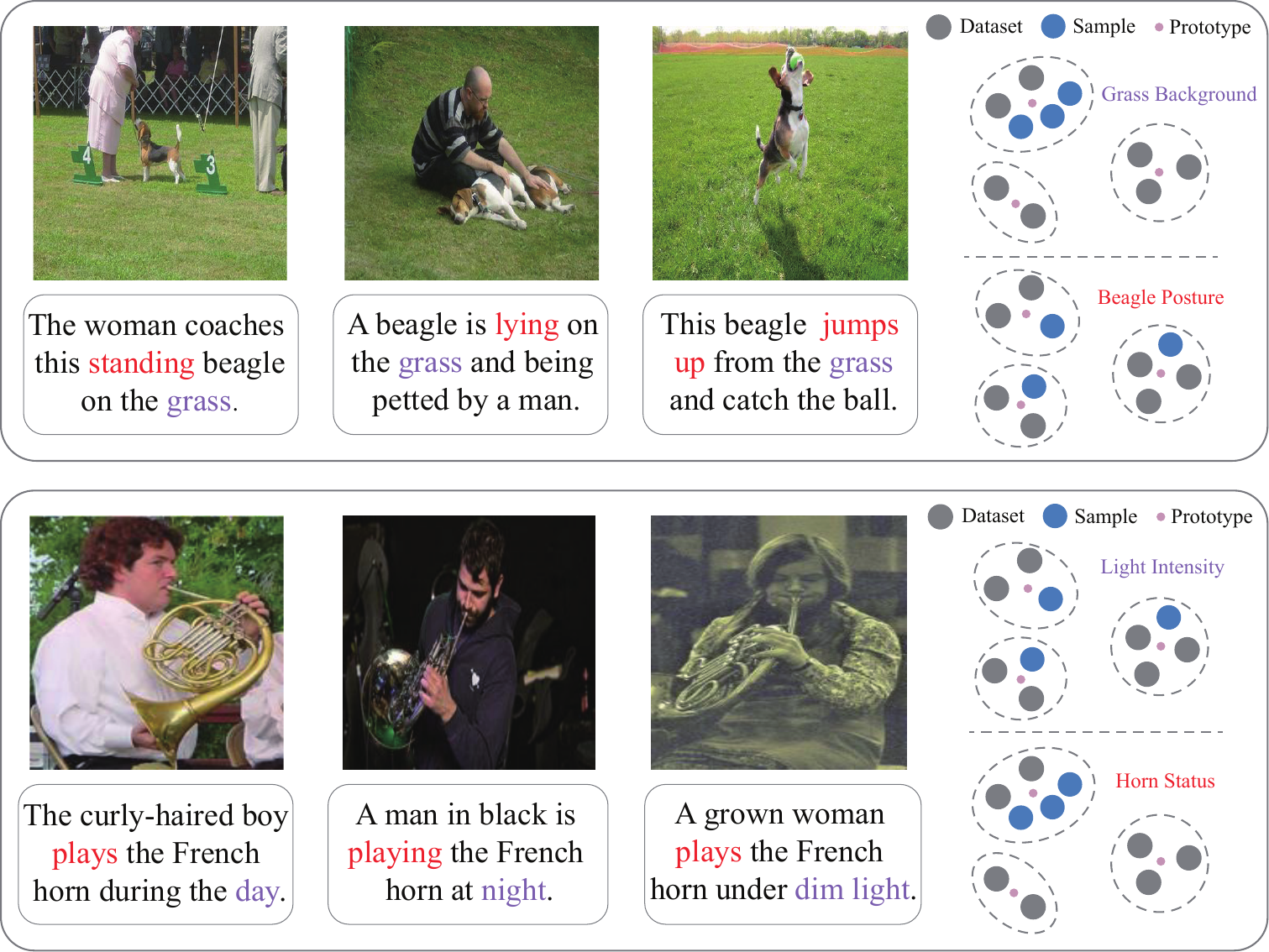}}
\caption{Classical DD relies only on weak visual information without leveraging valuable textual semantics. (Top) The uniform grassy background diminishes the salience of the beagle's posture. (Bottom) The light intensity affects the state expression of the played French horn.} 
\label{motivation}
\vspace{-5pt}
\end{figure}

\section{Introduction}
\label{sec:intro}
Driven by the exponential growth of intelligent systems and visual or image data, computer vision tasks have become increasingly critical across a wide range of domains \cite{cai2025computer, li2024learning}, while a core challenge remains the efficient learning of discriminative features from large-scale datasets. However, the acquisition, storage, and training processes for such datasets incur substantial computational costs and time overhead. To address this, \textit{Dataset Distillation} (DD) has emerged as a promising technique to compress the knowledge embedded within an original, large dataset into a compact set of synthetic samples. Models trained on this distilled dataset can achieve performance comparable to those trained on the original, full dataset, thereby significantly enhancing learning efficiency and reducing resource consumption.

In recent years, the matching-based DD approaches have consistently dominated, such as distribution matching \cite{zhao2023dataset}, gradient matching \cite{du2024diversity}, trajectory matching \cite{du2023minimizing,du2023sequential}, and dual-time matching \cite{yin2023squeeze}. However, they are usually computationally expensive and yield abstract, human-unreadable datasets. The lack of semantic interpretability also leads to poor performance in cross architecture generalization \cite{sun2024diversity}. Consequently, generation-based methods \cite{cazenavette2023generalizing} aimed at synthesizing more realistic images have arisen to address these limitations. Minimax \cite{gu2024efficient} introduces minimax criteria for diffusion models targeting the representativeness and diversity of generated data. D\textsuperscript{4}M \cite{su2024d} leverages the learned image prototypes to synthesize images via disentangled diffusion. Unfortunately, they rely solely on class or text labels to guide the diffusion process, meaning that all images of the same class share identical and scarce text caption during reverse sampling. This mode significantly compromises semantic richness and leads to substantial information loss \cite{li2023image}.

VLCP \cite{zou2025dataset} first proposes text prototypes to enhance the textual semantics of images in the distillation procedure. Nevertheless, text prototypes selected on the basis of word frequency suffer from poor interpretability, while image prototypes remain merely derived from visual features. As shown in Fig. \ref{motivation}, traditional learned image prototypes tend to focus on low-level visual textures such as background and illumination, failing to adequately leverage the implicit textual semantic inherent in images. This results in suboptimal prototypes and ultimately leads to performance degradation.

Building upon this insight, we propose \textbf{\underline{E}}nhancing Dataset \textbf{\underline{D}}istillation with \textbf{\underline{I}}mplicit \textbf{\underline{T}}extual \textbf{\underline{S}}emantics. We commence by utilizing a VLM (LLaVA \cite{liu2023visual}) to generate textual descriptions for the image dataset. These image-text pairs are then encoded into a joint feature space using CLIP \cite{radford2021learning}. \textbf{G}lobal \textbf{S}emantic \textbf{Q}uery (GSQ) module calculates the influence score of the entire text description on each image, fusing these scores to construct a preliminary cluster buffer. Subsequently, \textbf{L}ocal \textbf{S}emantic \textbf{A}wareness (LSA) module selects representative candidate samples from the buffer to facilitate prototype learning. The image prototypes are encoded and integrated via VAE encoder \cite{peebles2023scalable}, while the textual prototypes are summarized by an LLM (DeepSeek \cite{deepseekai2025deepseekr1incentivizingreasoningcapability}) guided by meticulously designed prompts. Finally, \textbf{D}ual-\textbf{P}rototype \textbf{G}uidance (DPG) synthesizes distilled dataset through the diffusion model.

Our main contributions are summarized as follows:
\begin{itemize}[itemsep=2pt, parsep=2pt, topsep=2pt, partopsep=3pt]
    \item We propose a novel DPG framework called EDITS that utilizes external textual signal to achieve enhanced DD.
    \item We introduce GSQ to comprehensively infuse textual semantics into image features, thereby strengthening the representational capacity of subsequent prototypes. 
    \item We employ LSA to identify representative candidate samples for the formation of image and text prototypes.
\end{itemize}

\section{methodology}
\label{sec:method}
\begin{figure*}[t!]
  \centering
  \centerline{\includegraphics[width=0.97\linewidth]{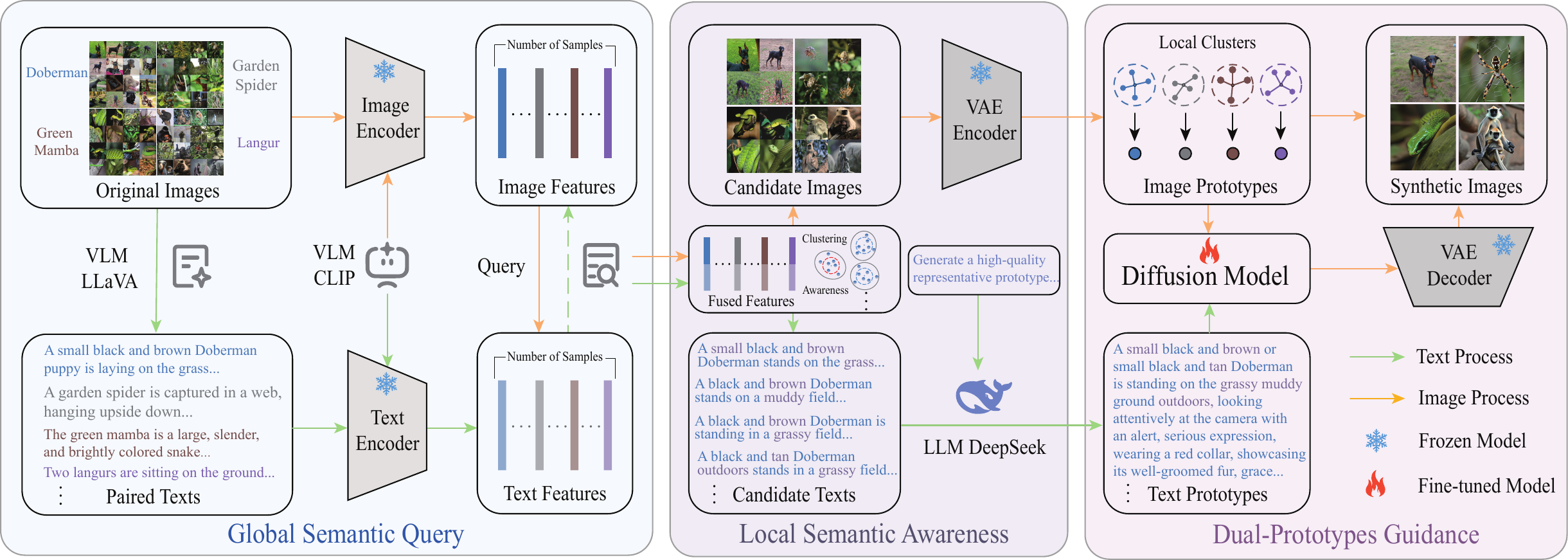}}
\caption{The overview of our EDITS. Global Semantic Query leverages texts generated by LLaVA to fuse image and text features. Local Semantic Awareness then selects representative candidate samples from the prior cluster buffer to construct image and text prototypes summarized by DeepSeek. Dual Prototype Guidance generates the synthetic dataset through a diffusion model.} 
\label{overview}
\end{figure*}

The overview of our EDITS is shown in Fig. \ref{overview}, including GSQ, LSA and DPG modules.

\subsection{Global Semantic Query}
Image datasets typically lack corresponding textual descriptions. To achieve our goal of semantic enhancement, we employ the open-source VLM (LLaVA \cite{liu2023visual}) to generate informative captions. This approach compensates for the absence of higher-level semantic information in visual features, such as logical relationships, causal chains, and temporal dynamics. The prompt is carefully designed to facilitate this process:

\vspace{3pt}
\begingroup
\ttfamily 
\noindent \seqsplit{Generate \ an \ extremely \ detailed \ and \ vivid \ caption \ for \ this \ image \ \{CLASS\} \dots}

\noindent Follow this structure: \dots
\endgroup
\vspace{3pt}



Consequently, we have successfully constructed a multi-modal image-text dataset $\mathcal{D}={\{({\bm{v}}_{i}, {\bm{\tau}}_{i}), {\bm{y}}_{i}\}}_{i=1}^{|\mathcal{D}|}$, where ${\bm{v}}_{i}$ and ${\bm{\tau}}_{i}$ are the $i$-th image and its paired caption, and ${\bm{y}}_{i} \in \mathcal{Y}=\{ 1, 2, \cdots , C\}$ is the corresponding ground-truth label, $C$ is the number of categories. $({\bm{v}}_{i}, {\bm{\tau}}_{i})\sim \mathcal{P}$, $\mathcal{P}$ is the natural data distribution. We use CLIP \cite{radford2021learning} with modality-specific encoders ${f}_{v}$ and ${f}_{\tau}$ to encode ${\bm{v}}_{i}, {\bm{\tau}}_{i}$ into a unified feature space, i.e., ${f}_{v}^{i}={f}_{v}({\bm{v}}_{i})$ and ${f}_{\tau}^{i}={f}_{\tau}({\bm{\tau}}_{i})$. For each image ${\bm{v}}_{i}$, we perform a global query over the textual semantics, computing influence scores from all textual descriptions:
\begin{equation}
{s}_{ij}=\frac{exp(I({f}_{v}^{i}, {f}_{\tau}^{j})/\mathcal{T})}{\textstyle\sum_{m=1}^{|\mathcal{D}|}exp(I({f}_{v}^{i}, {f}_{\tau}^{m})/\mathcal{T})}
\label{influence}
\end{equation}
\noindent where $I$ denotes the dot product, and $\mathcal{T}$ is the temperature.

The textual semantic features of the $i$-th image can be expressed as $\textstyle\sum_{j=1}^{|\mathcal{D}|}{s}_{ij}{f}_{\tau}^{j}$, the fusion feature is as follows:
\begin{equation}
{h}_{i}=concat(\textstyle\sum_{j=1}^{|\mathcal{D}|}{s}_{ij}{f}_{\tau}^{j}, {f}_{v}^{i})
\label{influence}
\end{equation}
\noindent where ${h}_{i}$ is the integration of visual-textual features for image $i$, providing superior expressive capacity.

Finally, we perform K-Means \cite{mcqueen1967some} clustering for \textbf{each category} based on 
$h$ to obtain cluster centers ${c}_{k}$ and their associated samples, which are stored in a buffer as priors for subsequent prototype selection. where $k\in\{1, 2, \cdots , K\}$, $K$=IPC (Image-Per-Class) is the number of cluster centers. 

\subsection{Local Semantic Awareness}
Unlike \cite{su2024d, zou2025dataset} where cluster centers serve directly as prototypes, our approach constructs enhanced prototypes through local semantic awareness around ${c}_{k}$, addressing its inherent semantic inadequacy. The awareness sample set of ${c}_{k}$ is:
\begin{equation}
{H}_{k}=\{(\bm{{v}}_{i}, {\bm{\tau}}_{i})|{h}_{i}\in\mathcal{R}_{{c}_{k}}\}
\label{influence}
\end{equation}
\noindent where $\mathcal{R}_{{c}_{k}}$ represents the local awareness radius.

The images from the awareness set are then processed by VAE encoder $\mathcal{E}$ to derive the \textbf{image prototype}:
\begin{equation}
{p}_{v}^{k}=\frac{\textstyle\sum_{i=1}^{|{H}_{k}|}\mathcal{E}({v}_{i})}{|{H}_{k}|}
\label{influence}
\end{equation}
The texts from the awareness set are summarized by the LLM (DeepSeek-Chat \cite{deepseekai2025deepseekr1incentivizingreasoningcapability}) into the \textbf{text prototype} ${p}_{\tau}^{k}$. The meticulously crafted prompt is:

\vspace{3pt}
\begingroup
\ttfamily 
\noindent \seqsplit{Please \ analyze\ the\ following\ \{len(TEXTS)\}\ texts\ and\ generate\ a\ high-quality\ repre-sentative\ prototype\ text.}

\noindent \#\# Input Texts: \{TEXTS\}

\noindent \#\# Output Requirements: 

\noindent \seqsplit{1. \ Extract \ semantic \ content\ directly\ re-lated\ to\ label\ \{CLASS\}\ in\ each\ text.}

\noindent \seqsplit{2. \ Merge\ unique\ information\ and\ express-ions\ from\ each\ text.}

\noindent \seqsplit{3. \ Fluent\ language,\ accurate\ information\ and\ clear\ structure \dots}
\endgroup
\vspace{3pt}

The rationale for summarizing only the texts within the awareness set, rather than all texts, lies in the following: (1) LLM inputs are subject to token limitations; (2) Excessive input increases the time overhead for generating text prototypes; (3) Incorporating excessive textual information may deviate from the semantic representation of the image prototype. Regarding point (3), our experiments indeed observe a performance degradation as the volume of text increased.

\subsection{Dual-Prototype Guidance}
Ultimately, image and text prototypes are combined within the Latent Diffusion Model \cite{rombach2022high} to generate diverse and representative images. The reverse process is:
\begin{equation}
\hat{{p}}_{t-1}^{\bm{v}}=s(\hat{{p}}_{t}^{\bm{v}}, t, {\epsilon}_{\theta }(\hat{{{p}}}_{t}^{\bm{v}}, t, {p}_{t}^{\bm{\tau}})) 
\label{influence}
\end{equation}
\noindent where $\hat{{p}}^{\bm{v}}$, ${p}^{\bm{\tau}}$ is the image prototype with noise and text prototype, ${\epsilon}_{\theta }$ is the noise predictor (U-Net), and $s$ is the scheduler. The synthetic dataset is $\mathcal{S}=\{\mathcal{F}(\hat{{p}}_{0}^{\bm{v}})\}$. $\mathcal{F}$ is the VAE decoder.
\section{Experiments}
\label{sec:exp}
\subsection{Experimental Setup}
\textbf{Datasets}. Experiments are conducted on ImageNet subsets (256$\times$256) \cite{howard2019smaller, kim2022dataset} , CIFAR10 and CIFAR100 (32$\times$32).

\vspace{3pt}
\noindent \textbf{Baselines and evaluation.} We compare EDITS with some state-of-the-art coreset selection and DD methods, including K-Center \cite{mcqueen1967some}, Herding \cite{welling2009herding}, raw DiT \cite{peebles2023scalable}, DM \cite{zhao2023dataset}, IDC \cite{kim2022dataset}, Minimax \cite{gu2024efficient}, D\textsuperscript{4}M \cite{su2024d}, VLCP \cite{zou2025dataset}, $\text{SRe}^\text{2}\text{L}$ \cite{yin2023squeeze}, and RDED \cite{sun2024diversity} with hard-label \cite{gu2024efficient} and soft-label evaluation \cite{sun2024diversity}. 

\vspace{3pt}
\noindent \textbf{Implementation details.} The image-text pairs are used to fine-tune the Stable 
Diffusion V1-5 \cite{rombach2022high} individually for each dataset. We use $\textit{DDIM}$ as the scheduler and $\textit{vae-ft-mse}$ as the VAE model. The capacity of the awareness set is 5 samples.

\subsection{Main Results}
\textbf{ImageNet Subsets.} We conduct extensive experiments under various IPC and architecture settings on ImageWoof, ImageNette, and ImageIDC. The results in Tab. \ref{nette_idc} and Tab. \ref{woof} show that EDITS consistently outperforms existing state-of-the-art methods by about 1\%-3\% on high-resolution datasets, which fully demonstrates its robustness and generalization across different datasets and architectures. We observe that EDITS achieves remarkable performance on the fine-grained ImageWoof, which can be attributed to the discriminative textual semantics provided by the prior knowledge of the VLM.

\begin{table}[t!]
\centering
\caption{Comparison of various methods on ImageNette and ImageIDC. All the results are obtained on ResNetAP-10.}
\resizebox{\linewidth}{!}{%
\begin{tabular}{lccccccc}
\toprule
Dataset & IPC & Random & DiT                                   & Minimax & D\textsuperscript{4}M  & VLCP                           & \textbf{EDITS}                       \\ \midrule
 & 10 & 54.2\scriptsize{±1.6} & 59.1\scriptsize{±0.7}                             & 57.7\scriptsize{±1.2} &60.9\scriptsize{±1.7} & \underline{61.3\scriptsize{±0.5}}   & \textbf{64.5\scriptsize{±0.1}}              \\
 Nette & 20 & 63.5\scriptsize{±0.5} & 64.8\scriptsize{±1.2}              & 64.7\scriptsize{±0.8} & 66.3\scriptsize{±1.3} & \underline{69.6\scriptsize{±0.3}}               & \textbf{71.3\scriptsize{±0.3}}    \\
  & 50 & 76.1\scriptsize{±1.1} & 73.3\scriptsize{±0.9}                            & 73.9\scriptsize{±0.3} & 77.7\scriptsize{±1.1} & \underline{77.8\scriptsize{±0.5}}   & \textbf{78.5\scriptsize{±0.3}}               \\ \midrule
 & 10 & 48.1\scriptsize{±0.8} & 54.1\scriptsize{±0.4}                 & 51.9\scriptsize{±1.4} &50.3\scriptsize{±1.0} & \underline{55.7\scriptsize{±1.4}}                & \textbf{56.5\scriptsize{±1.3}}                \\
 IDC & 20 & 52.5\scriptsize{±0.9} & 58.9\scriptsize{±0.2}                & 59.1\scriptsize{±3.7}  & 55.8\scriptsize{±0.2} & \underline{60.4\scriptsize{±0.8}}  & \textbf{61.5\scriptsize{±0.6}}                \\
  & 50 & 68.1\scriptsize{±0.7} & 64.3\scriptsize{±0.6}                            & 69.4\scriptsize{±1.4} & 69.1\scriptsize{±2.4} & \underline{70.7\scriptsize{±1.1}}   & \textbf{71.8\scriptsize{±0.5}}                \\ \bottomrule
\end{tabular}%
}
\label{nette_idc}
\end{table}


\begin{table}[t!]
\centering
\scriptsize
\caption{The results on CIFAR-10 and CIFAR-100. We use ResNet-18 to retrieve the distilled data and evaluate. }
\setlength{\arrayrulewidth}{0.48pt}
\renewcommand{\arraystretch}{1.1}

\resizebox{\linewidth}{!}{%
\begin{tabular}{lcccccc}
\hline
Dataset & IPC & $\text{SRe}^\text{2}\text{L}$  & RDED   & VLCP & \textbf{EDITS} \\ \hline
\multirow{2}{*}{CIFAR-10}  & 10  & 29.3\scriptsize{±0.5} &  37.1\scriptsize{±0.3}  & \underline{39.0\scriptsize{±0.7}}  & \textbf{42.1\scriptsize{±0.5}}  \\
                            & 50  & 45.0\scriptsize{±0.7} & 62.1\scriptsize{±0.1} & \underline{63.2\scriptsize{±0.3}}  & \textbf{64.5\scriptsize{±0.2}}  \\ \hline
\multirow{2}{*}{CIFAR-100} & 10  & 27.0\scriptsize{±0.4} &  42.6\scriptsize{±0.2}  & \underline{50.6\scriptsize{±0.7}}  & \textbf{52.8\scriptsize{±0.3}}   \\
                            & 50  & 50.2\scriptsize{±0.4} & 62.6\scriptsize{±0.1} & \underline{66.1\scriptsize{±0.3}}  & \textbf{68.0\scriptsize{±0.7}}   \\ \hline
\end{tabular}
}
\label{cifar}
\end{table}

\vspace{3pt}
\noindent \textbf{CIFAR10 and CIFAR100.} As illustrated in Tab. \ref{cifar}, we also evaluate the performance of EDITS on the low-resolution CIFAR10 and CIFAR100. EDITS surpasses VLCP, which confirms that our comprehensive semantic enhancement is more effective than the method that only utilize text prototypes.


\begin{table*}[t!]
\centering
\caption{Performance comparison of state-of-the-art methods on ImageWoof with different IPC (compression ratio) and model architectures. All the results are obtained on the 256 $\times$ 256 resolution. The results of VLCP are reproduced under identical settings. The best and second best results are emphasized in \textbf{bold} and \underline{underlined} cases.}
\setlength{\arrayrulewidth}{0.8pt}
\resizebox{\textwidth}{!}{%
\begin{tabular}{llccccccccccc}
\toprule
IPC (Ratio) & Evaluate Model & Random & K-Center & Herding & DiT  & DM  & IDC-1  & Minimax & D\textsuperscript{4}M  & VLCP  & \textbf{EDITS} & Full \\ \midrule 
 & ConvNet-6 & 24.3\scriptsize{±1.1} & 19.4\scriptsize{±0.9} & 26.7\scriptsize{±0.5} & \underline{34.2\scriptsize{±1.1}} & 26.9\scriptsize{±1.2} & 33.3\scriptsize{±1.1}  & 33.3\scriptsize{±1.7} & 29.4\scriptsize{±0.9} & 33.8\scriptsize{±0.4} & \textbf{34.6\scriptsize{±0.4}} & 86.4\scriptsize{±0.2} \\
 10 (0.8\%) & ResNetAP-10 & 29.4\scriptsize{±0.8} & 22.1\scriptsize{±0.1} & 32.0\scriptsize{±0.3} & 34.7\scriptsize{±0.5} & 30.3\scriptsize{±1.2} & \underline{39.1\scriptsize{±0.5}}  & 36.2\scriptsize{±3.2} & 33.2\scriptsize{±2.1} & 37.7\scriptsize{±0.3} & \textbf{39.4\scriptsize{±0.5}} & 87.5\scriptsize{±0.5} \\
   & ResNet-18 & 27.7\scriptsize{±0.9} & 21.1\scriptsize{±0.4} & 30.2\scriptsize{±1.2} & 34.7\scriptsize{±0.4} & 33.4\scriptsize{±0.7} & \underline{37.3\scriptsize{±0.2}}  & 35.7\scriptsize{±1.6} & 32.3\scriptsize{±1.2} & 36.1\scriptsize{±0.1} & \textbf{38.1\scriptsize{±1.3}} & 89.3\scriptsize{±1.2} \\ \midrule
 & ConvNet-6 & 29.1\scriptsize{±0.7} & 21.5\scriptsize{±0.8} & 29.5\scriptsize{±0.3} & 36.1\scriptsize{±0.8} & 29.9\scriptsize{±1.0} & 35.5\scriptsize{±0.8}  & \underline{37.3\scriptsize{±0.1}} & 34.0\scriptsize{±2.3} & 35.1\scriptsize{±0.5} & \textbf{39.3\scriptsize{±0.3}} & 86.4\scriptsize{±0.2} \\
  20 (1.6\%) & ResNetAP-10 & 32.7\scriptsize{±0.4} & 25.1\scriptsize{±0.7} & 34.9\scriptsize{±0.1} & 41.1\scriptsize{±0.8} & 35.2\scriptsize{±0.6} & \underline{43.4\scriptsize{±0.3}}  & 43.3\scriptsize{±2.7} & 40.1\scriptsize{±1.6} & 41.3\scriptsize{±0.2} & \textbf{43.7\scriptsize{±1.1}} & 87.5\scriptsize{±0.5} \\
   & ResNet-18 & 29.7\scriptsize{±0.5} & 23.6\scriptsize{±0.3} & 32.2\scriptsize{±0.6} & 40.5\scriptsize{±0.5} & 29.8\scriptsize{±1.7} & 38.6\scriptsize{±0.2}  & 41.8\scriptsize{±1.9} & 38.4\scriptsize{±1.1} & \underline{41.9\scriptsize{±1.4}} & \textbf{44.1\scriptsize{±0.7}} & 89.3\scriptsize{±1.2} \\ \midrule
 & ConvNet-6 & 41.3\scriptsize{±0.6} & 36.5\scriptsize{±1.0} & 40.3\scriptsize{±0.7} & 46.5\scriptsize{±0.8} & 44.4\scriptsize{±1.0} & 43.9\scriptsize{±1.2}  & \underline{50.9\scriptsize{±0.8}} & 47.4\scriptsize{±0.9} & 50.3\scriptsize{±1.1} & \textbf{51.4\scriptsize{±0.6}} & 86.4\scriptsize{±0.2} \\
  50 (3.8\%) & ResNetAP-10 & 47.2\scriptsize{±1.3} & 40.6\scriptsize{±0.4} & 49.1\scriptsize{±0.7} & 49.3\scriptsize{±0.2} & 47.1\scriptsize{±1.1} & 48.3\scriptsize{±1.0}  & 53.9\scriptsize{±0.7} & 51.7\scriptsize{±3.2} & \underline{55.5\scriptsize{±1.2}} & \textbf{56.1\scriptsize{±0.7}} & 87.5\scriptsize{±0.5} \\
   & ResNet-18 & 47.9\scriptsize{±1.8} & 39.6\scriptsize{±1.0} & 48.3\scriptsize{±1.2} & 50.1\scriptsize{±0.5} & 46.2\scriptsize{±0.6} & 48.3\scriptsize{±0.8} & 53.7\scriptsize{±0.6} & 53.7\scriptsize{±2.2} & \underline{56.2\scriptsize{±0.4}} & \textbf{56.8\scriptsize{±0.6}}  & 89.3\scriptsize{±1.2} \\ \midrule
 & ConvNet-6 & 46.3\scriptsize{±0.6} & 38.6\scriptsize{±0.7} & 46.2\scriptsize{±0.6} & 50.1\scriptsize{±1.2} & 47.5\scriptsize{±0.8} & 48.9\scriptsize{±0.7}  & 51.3\scriptsize{±0.6} & 50.5\scriptsize{±0.4} & \underline{53.3\scriptsize{±1.3}} & \textbf{55.8\scriptsize{±0.6}} & 86.4\scriptsize{±0.2} \\
  70 (5.4\%) & ResNetAP-10 & 50.8\scriptsize{±0.6} & 45.9\scriptsize{±1.5} & 53.4\scriptsize{±1.4} & 54.3\scriptsize{±0.9} & 51.7\scriptsize{±0.8} & 52.8\scriptsize{±1.8} & 57.0\scriptsize{±0.2} & 54.7\scriptsize{±1.6} & \underline{57.5\scriptsize{±0.2}}  & \textbf{59.7\scriptsize{±1.0}} & 87.5\scriptsize{±0.5} \\
   & ResNet-18 & 52.1\scriptsize{±1.0} & 44.6\scriptsize{±1.1} & 49.7\scriptsize{±0.8} & 51.5\scriptsize{±1.0} & 51.9\scriptsize{±0.8} & 51.1\scriptsize{±1.7} & 56.5\scriptsize{±0.8} & 56.3\scriptsize{±1.8} & \underline{57.3\scriptsize{±0.5}} & \textbf{58.4\scriptsize{±1.3}}  & 89.3\scriptsize{±1.2} \\ \midrule
 & ConvNet-6 & 52.2\scriptsize{±0.4} & 45.1\scriptsize{±0.5} & 54.4\scriptsize{±1.1} & 53.4\scriptsize{±0.3} & 55.0\scriptsize{±1.3} & 53.2\scriptsize{±0.9}  & 57.8\scriptsize{±0.9} & 57.9\scriptsize{±1.5} & \underline{58.7\scriptsize{±1.1}} & \textbf{59.3\scriptsize{±1.4}} & 86.4\scriptsize{±0.2} \\
 100 (7.7\%) & ResNetAP-10 & 59.4\scriptsize{±1.0} & 54.8\scriptsize{±0.2} & 61.7\scriptsize{±0.9} & 58.3\scriptsize{±0.8} & 56.4\scriptsize{±0.8} & 56.1\scriptsize{±0.9}  & 62.7\scriptsize{±1.4} & 59.5\scriptsize{±1.8} & \underline{63.3\scriptsize{±0.5}} & \textbf{64.4\scriptsize{±1.2}} & 87.5\scriptsize{±0.5} \\
   & ResNet-18 & 61.5\scriptsize{±1.3} & 50.4\scriptsize{±0.4} & 59.3\scriptsize{±0.7} & 58.9\scriptsize{±1.3} & 60.2\scriptsize{±1.0} & 58.3\scriptsize{±1.2}  & 62.7\scriptsize{±0.4} & 63.8\scriptsize{±1.3} & \underline{65.2\scriptsize{±0.8}} & \textbf{66.3\scriptsize{±1.4}} & 89.3\scriptsize{±1.2} \\
\bottomrule
\end{tabular}%
}
\label{woof}
\end{table*}

\begin{table}[t!]
\centering
\caption{The isolated and combinatorial effects of our proposed modules. w/o DPG means using only text labels instead of text prototypes and image prototypes for guidance.}
\renewcommand{\arraystretch}{1.0}
\small
\setlength{\tabcolsep}{5.0pt} 
\begin{tabular}{cccccc}
\hline
GSQ           & LSA            & DPG            & ConvNet-6               & ResNetAP-10             & ResNet-18            \\    \hline
 $\times$     & $\times$       & $\times$       & 60.7\scriptsize{±1.6}   & 61.0\scriptsize{±0.6}   & 62.5\scriptsize{±0.7}      \\ 
 $\times$     &  $\checkmark$  & $\times$       & 61.4\scriptsize{±1.3}   & 63.2\scriptsize{±0.7}   & 64.4\scriptsize{±0.5}     \\ 
 $\checkmark$ & $\times$       & $\times$       & 61.6\scriptsize{±0.6}   & 62.5\scriptsize{±1.2}   & 63.1\scriptsize{±1.5}    \\
 $\checkmark$ & $\checkmark$   & $\times$       & 62.0\scriptsize{±0.9}   & 63.8\scriptsize{±1.1}   & 65.2\scriptsize{±0.4}    \\
 $\checkmark$ & $\checkmark$   & $\checkmark$   & \textbf{62.9\scriptsize{±1.0}}   & \textbf{64.5\scriptsize{±0.1}}   & \textbf{66.7\scriptsize{±0.3}}    \\\hline
\end{tabular}
\label{component}
\end{table}

\begin{figure}[t]
    \centering
    \begin{subfigure}[b]{0.47\columnwidth} 
        \centering
        \includegraphics[width=\textwidth]{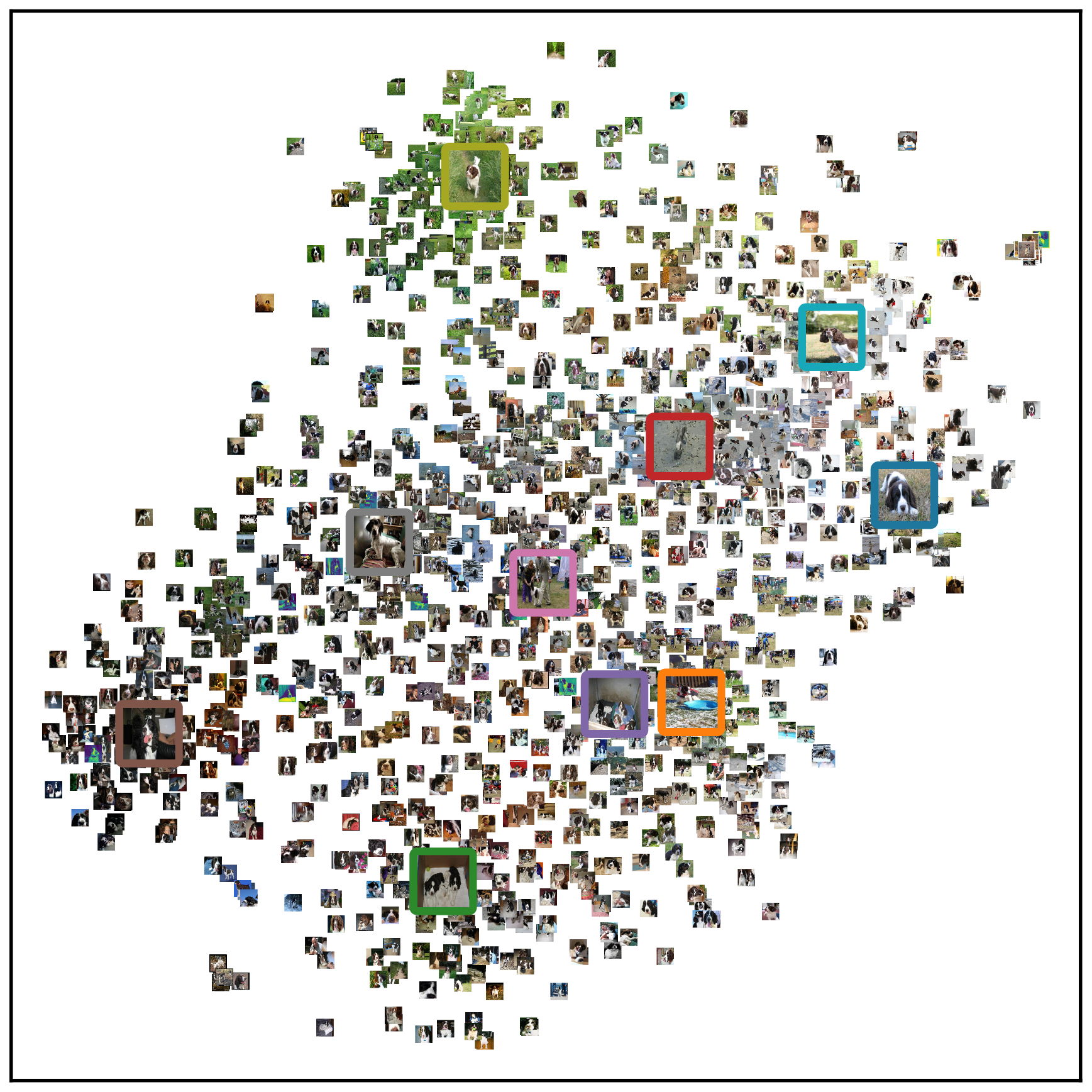}
        \caption{Vanilla}
    \end{subfigure}
    \hfill
    \begin{subfigure}[b]{0.47\columnwidth} 
        \centering
        \includegraphics[width=\textwidth]{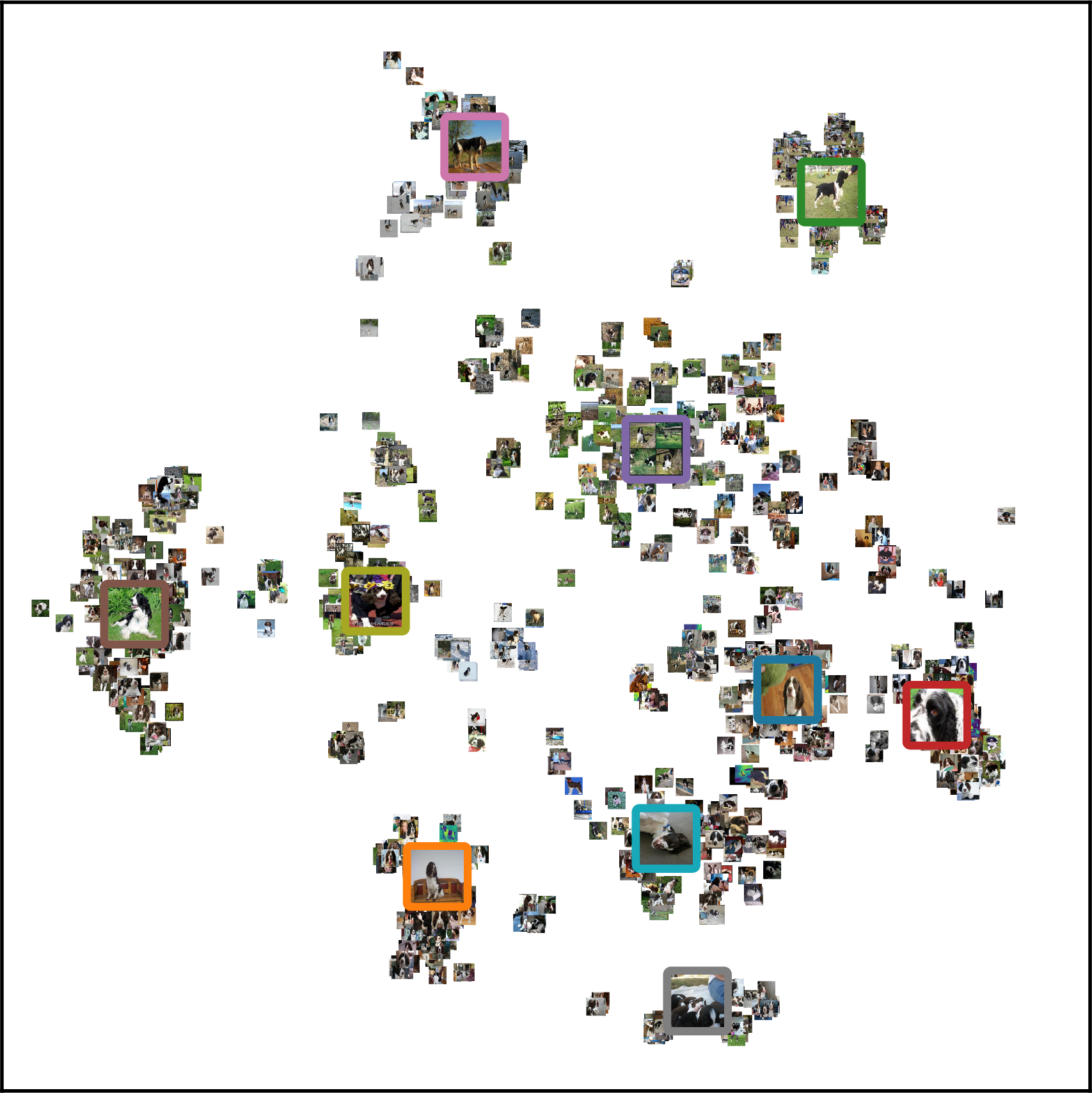}
        \caption{Ours}
    \end{subfigure}
    \caption{The t-SNE visualization contrast the image prototypes from the traditional method (left) with those from our EDITS (right) under IPC=10. The label is ``English springer". } 
    \label{compare_tsne}
\end{figure}


\subsection{Ablation Studies}
\noindent \textbf{Effect of Each Module.} As shown in Tab. \ref{component}, we incrementally evaluate the impact of each proposed modules on ImageNette (IPC=10).  Every individual component plays a critical role in determining and enhancing the system's overall performance. When all components are combined, our EDITS achieves its optimal performance, resulting in a substantial improvement (e.g., a 4.2\% gain on ResNet-18). This fully demonstrates that our external semantic guidance is effective.

\noindent \textbf{Visualization.} Our text prototypes are necessarily more semantically rich than ordinary text labels. And we visualize the comparison between the image prototype that incorporates external text semantics and the vanilla prototype in Fig. \ref{compare_tsne}. The vanilla approach performs prototype learning in the latent space of VAE-encoded image features, focusing on low-level texture attributes such as background patterns. For example, it clusters images with grassy backgrounds in the upper-left region, resulting in minimal inter-class separation and poor discriminability. In contrast, our method places greater emphasis on semantic representation. For instance, in the upper-right region, it clusters images of dogs interacting with humans, while maintaining greater inter-class dispersion.

\section{Conclusion}
\label{sec:con}
In this work, we introduce EDITS, a novel DD framework that leverages implicit textual semantics to enhance the representational capacity of images. By integrating Global Semantic Query for fusing textual and visual features, Local Semantic Awareness for prototype construction, and Dual-Prototype Guidance for diffusion-based generation, our method effectively addresses the limitations of traditional approaches that rely solely on low-level visual cues or limited text labels.
\vspace{2pt}

\noindent \textbf{Future Works.} Beyond achieving higher classification accuracy, EDITS highlights the importance of incorporating rich semantic into image dataset distillation, offering a more interpretable and robust paradigm for future research. Moving forward, we plan to explore scaling EDITS to larger datasets, improving efficiency in low-resource settings, and extending the framework to multi-modal or domain-specific applications such as medical imaging and autonomous driving.
\vfill\pagebreak

\bibliographystyle{IEEEbib}
\bibliography{strings,refs}

\end{document}